\documentclass{article}

\usepackage[numbers]{natbib}

    \usepackage[preprint]{EMoG}

\usepackage[utf8]{inputenc} 
\usepackage[T1]{fontenc}    
\usepackage{hyperref}       
\usepackage{url}            
\usepackage{booktabs}       
\usepackage{amsfonts}       
\usepackage{nicefrac}       
\usepackage{microtype}      
\usepackage{xcolor}         
\usepackage{caption}
\usepackage{graphicx, subfig}
\usepackage{bm}

\usepackage{floatrow}
\newfloatcommand{capbtabbox}{table}[][\FBwidth]

\title{EMoG: Synthesizing Emotive Co-speech 3D Gesture with Diffusion Model}

\author{%
  Lianying Yin$^{1,2}$\footnotemark[1] ,  Yijun Wang$^{1}$\footnotemark[1] ,  Tianyu He$^{3}$\footnotemark[1] , Jinming Liu$^{2,4}$,  Wei Zhao$^{1,2}$,  \\
  \textbf{Bohan Li}$^{2,4}$\textbf{,} \textbf{Xin Jin}$^{2}$\footnotemark[2] \textbf{,} \textbf{Jianxin Lin}$^{1}$\footnotemark[2] \\
$^{1}$ The College of Computer Science and Electronic Engineering, Hunan University \\
$^{2}$ Eastern Institute for Advanced Study\\
$^{3}$ Microsoft Research Asia\\
$^{4}$ Shanghai Jiao Tong University\\
{\tt\small \{yin2110, wyjun, zhaoweiheap, linjianxin\}@hnu.edu.cn} \\
{\tt\small \{deeptimhe, bohan.li77\}@gmail.com, jmliu206@sjtu.edu.cn, jinxin@eias.ac.cn}\\
}

\begin{document}
\maketitle

\renewcommand{\thefootnote}{\fnsymbol{footnote}}
\footnotetext[1]{Equal contribution.}
\footnotetext[2]{Corresponding author.}

\begin{abstract}
  Although previous co-speech gesture generation methods are able to synthesize motions in line with speech content, it is still not enough to handle diverse and complicated motion distribution. The key challenges are: 1) the one-to-many nature between the speech content and gestures; 2) the correlation modeling between the body joints. In this paper, we present a novel framework (EMoG) to tackle the above challenges with denoising diffusion models: 1) To alleviate the one-to-many problem, we incorporate emotion clues to guide the generation process, making the generation much easier; 2) To model joint correlation, we propose to decompose the difficult gesture generation into two sub-problems: joint correlation modeling and temporal dynamics modeling. Then, the two sub-problems are explicitly tackled with our proposed Joint Correlation-aware transFormer (JCFormer). Through extensive evaluations, we demonstrate that our proposed method surpasses previous state-of-the-art approaches, offering substantial superiority in gesture synthesis.
\end{abstract}

\section{Introduction}

Aside from speech content, co-speech gesture conveys additional information like personality, and emotion in human communication~\cite{cassell1999speech,mcneill1992hand,wagner2014gesture}. It also contributes to a more engaging and interactive exchange between speakers and listeners, fostering a greater sense of connection and understanding~\cite{burgoon1990nonverbal,butterworth1989gesture,huang2012robot}. Therefore, co-speech gesture generation has gained significant attention in recent years due to its potential applications in various fields, such as films, games, and avatar animation.


Early methods tackle the co-speech gesture generation with deterministic models (e.g., Convolution Neural Networks, Recurrent Neural Networks) \cite{habibie2021learning,bhattacharya2021speech2affectivegestures,liu2022learning,yoon2019robots,yoon2020speech} tend to produce dull average results. Considering the inherent one-to-many relationship between audio and gesture, generative models, such as flow-based models, VAEs, provide an alternative approach to generating gesture motion from audio \cite{alexanderson2020style,ye2022audio,ghorbani2023zeroeggs,li2021audio2gestures,liu2022audio,yazdian2022gesture2vec,yi2022generating}. Despite recent works using Generative Adversarial Networks (GANs) could ensure realistic results to some extent, the issues of mode collapse and unstable training still hinder the ability to achieve high-fidelity gesture synthesis conditioned on audio. Recently, some studies~\cite{zhu2023taming} begin to explore the diffusion model to establish multi-modal correspondences, achieving state-of-the-art co-speech gesture generation performance.

While significant progress has been made in matching audio to generated gestures \cite{ginosar2019learning,yoon2020speech,li2021audio2gestures,zhu2023taming}, few studies explore emotive gesture generation. CaMN \cite{liu2022beat} incorporates emotion labels as additional input to regulate emotional expression. Meanwhile, concurrent work \cite{ao2023gesturediffuclip} adapts CLIP's \cite{radford2021learning} latent space to control emotion through text prompts. All these techniques aim at generating emotive co-speech gestures that require additional inputs as regularization, as the complex mapping from audio to gesture often causes loss of non-critical properties such as subtle emotions, which makes it hard to directly generate emotive gestures from audio. Meanwhile, existing methods mainly use joint coordinates to form feature vectors at each frame and perform temporal analysis, but their ability is limited as they do not explicitly utilize the spatial relationships between human joints, resulting in suboptimal generated gestures.

To address the above challenges, we propose a novel audio-driven diffusion-based framework, dubbed \textbf{EMo}tive \textbf{G}esture generation (EMoG), to generate vivid co-speech gestures. The core insight is to extract emotional attributes from audio, use diffusion models to finely model the joint distribution of audio and gesture, and further decouple the spatial-joint relationships and temporal dynamics of gesture to achieve high-fidelity gesture reconstruction. 
Specifically, we first utilize an audio representation obtained from the pre-trained wav2vec 2.0 model \cite{baevski2020wav2vec} to extract the emotion attribute which serves as a crucial control factor in generating emotive gestures, using a supervised manner. Then, finding an optimal way to integrate emotion into gesture is critical for achieving optimal performance. We elaborate on the relationship between emotion and gesture sequences, providing a detailed exploration of methods for integrating emotion: 

1) We envision a global relationship between emotion modality and gesture sequences. We employ the Adaptive Layer Normalization (AdaLN) \cite{perez2018film} which modulates the channel-map distribution to apply the emotion modality to the entire gesture sequence. We also explore the in-context initialization method which is another method for establishing global connections and involves augmenting the noisy gesture sequence with the emotion attribute globally. 

2) Another envision is that emotions are subject to temporal change, with their intensity varying from moment to moment. We utilize a commonly used cross-attention approach \cite{rombach2022high} which uses the emotional vector to compute an attention score that modulates the vivid gesture generation at each frame.

To address the second challenge, our solution is proposed based on the fact that spatial information, which encodes kinematic relationships between joints and correlations between different body parts, is as vital as temporal information. We carefully design a multi-modality Joint Correlation-aware transFormer (JCFormer) module to fuse the spatial-joint and temporal information. The spatial-joint correlations extracted by the joint-aware transformer are broadcast to the temporal dynamic gesture obtained by the temporal-aware transformer for further fusion.

Our main contributions can be summarized as three-tiered: 1) As an early attempt to extract emotions from audio and generate emotive co-speech gestures using the diffusion-based model, we explored various possibilities of integrating emotions into gestures, resulting in the synthesis of vivid co-speech gestures. 
2) We propose a Joint Correlation-aware Transformer to disentangle the gestures to spatial-joint correlations and inter-frame temporal dynamics. 3) Extensive experiments demonstrate that the proposed framework generates vivid and realistic co-speech gestures.





\section{Related Work}

\paragraph{Co-speech Gesture Synthesis}

The advanced modeling capabilities of deep neural networks enable the training of sophisticated end-to-end models, utilizing raw speech-gesture data directly. One option is deterministic models, such as MLP \cite{kucherenko2020gesticulator}, CNN \cite{habibie2021learning}, RNN \cite{bhattacharya2021speech2affectivegestures, liu2022learning,yoon2019robots,yoon2020speech}, and Transformer \cite{bhattacharya2021text2gestures}. The inherent one-to-many relationship between audio and gesture motion often results in deterministic models that generate uninteresting average poses. Generative models, including flow-based models, VAEs, and VQ-VAE, offer an alternative approach to generate gesture motion from audio \cite{alexanderson2020style,ye2022audio,ghorbani2023zeroeggs,li2021audio2gestures,liu2022audio,yazdian2022gesture2vec,yi2022generating}. Despite the use of GANs in recent works to ensure realistic results, the well-known issues of mode collapse and unstable training impede the ability to achieve high-fidelity gesture distribution learning conditioned on audio. Recently, diffusion models have demonstrated remarkable performance in image synthesis and human motion generation tasks \cite{ho2020denoising,sohl2015deep,song2020improved,ramesh2022hierarchical,tevet2022human,zhang2022motiondiffuse}. There are some works that have been explored in this area. For instance, \cite{ao2023gesturediffuclip} introduced a gesture diffusion model that incorporated CLIP latents for improved gesture generation. \cite{zhu2023taming} proposed a diffusion-based framework called DiffGesture for high-fidelity audio-driven co-speech gesture generation, which used a Diffusion Audio-Gesture Transformer and a Diffusion Gesture Stabilizer to capture cross-modal associations and preserve temporal coherence, achieving state-of-the-art performance. Since DiffGesture \cite{zhu2023taming} only focuses on temporal dynamics of motion, they have a great performance on simple 3D keypoints representation but failed in generating high-fidelity gestures using complex relative-skeletal space representation. Different from DiffGesture\cite{zhu2023taming}, our approach exploits the interdependence between joints to generate more optimal co-speech gestures.
\paragraph{Emotive Gesture Synthesis}

Currently, there is limited research on generating emotive human motion based on audio input. Previous approaches have involved using emotion labels alongside other input modalities to generate emotive human gestures. For example, \cite{liu2022beat} concatenated emotion labels into other input modalities, another concurrent work \cite{ao2023gesturediffuclip} adapted the CLIP's latent space from text and image modalities to text and gesture modalities using contrastive learning. In contrast, our method extracts emotional features directly from audio and integrates them into the distributional modeling process of the diffusion model. This results in high-quality emotive human gestures without deriving emotion from additional modal input.


\begin{figure*}[t]
	\centering
	\includegraphics[width=1.1\textwidth]{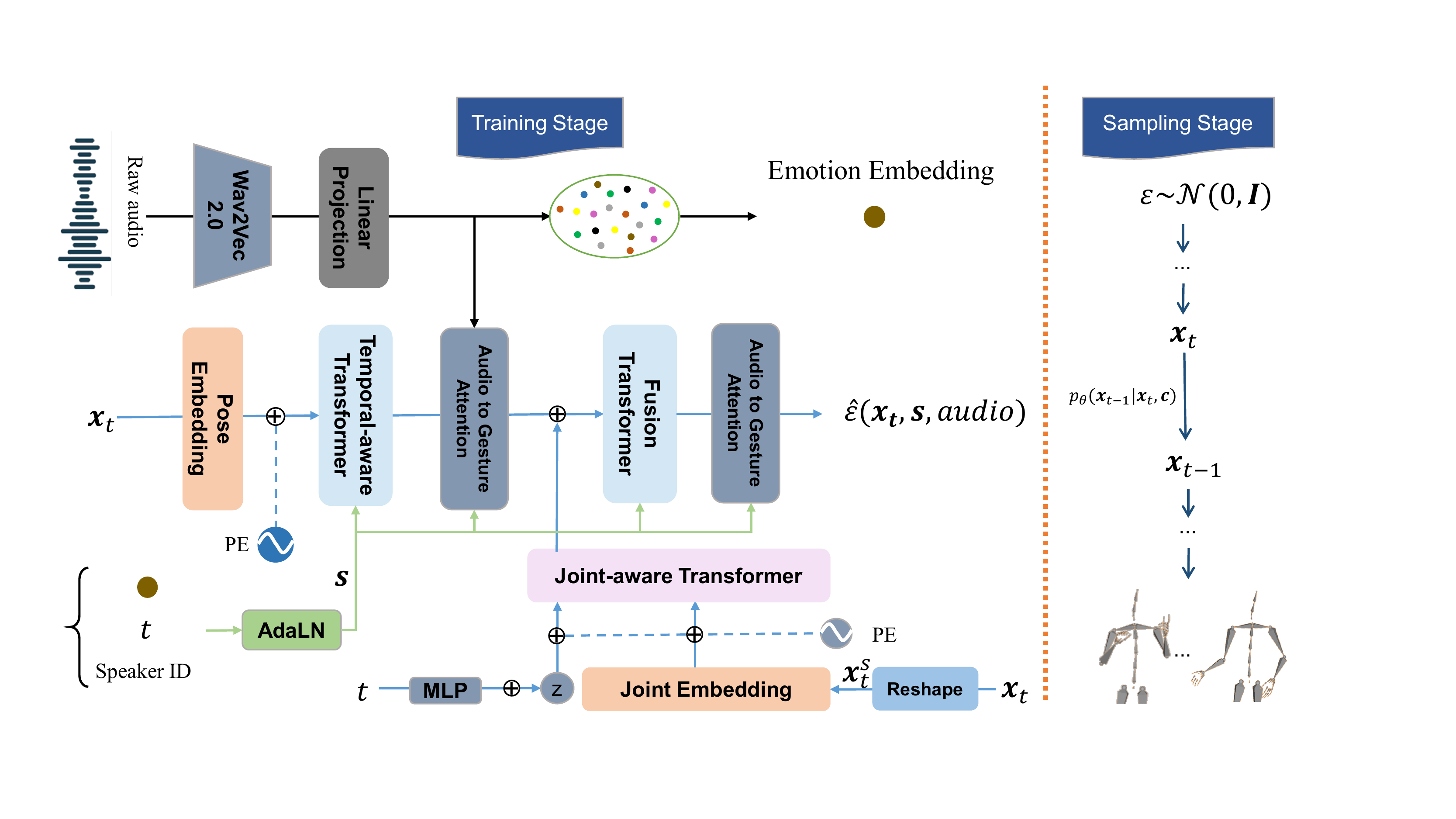}
	\caption{{Overview of the EMotive co-speech 3D Gesture synthesizing (EMoG) framework.}} 
	\label{fig:network}
\end{figure*}
\section{Methodology}\label{Methodology}
We present an EMotive co-speech 3D Gesture synthesizing framework (EMoG) that aims to achieve free gesture control driven by audio signals and emotion clues as shown in Figure \ref{fig:network}. We introduce an audio-conditioned gesture generation pipeline using a denoising diffusion model in Section \ref{sec:diff}. We further elaborate on the relationship between emotion and gesture sequence in Section \ref{sec:ieig}, providing a detailed exploration of the methods for integrating emotions. Finally, we delicately design a Joint Correlation-aware transFormer (JCFormer) to predict the joints' correlation and dynamics simultaneously and illustrate it in Section \ref{sec:trans}.

\subsection{Problem Definition}
\label{Problem Definition}
The main objective of this paper is to generate a gesture sequence $\bm{x}=x^{1:N}$, where $N$ represents the total frames, $x^{i} \in \mathbb{R}^{J\times3}$ represents the 3D rotation pose state in the $i$-th frame, based on a corresponding sequence of audio features $\bm{A} = a^{1:N}$, and optionally an emotional vector $\bm{e}$ extracted from $\bm{A}$, where $J$ is the number of joints.

\subsection{Diffusion Model for Gesture Synthesis}\label{sec:diff}  
A probabilistic model is trained to gradually remove noise from a Gaussian distribution to generate a desired output, and generally consists of a forward diffusion process and a reverse process.

\paragraph{The Forward Diffusion Process.} To approximate the posterior $q({x}_{1:T}^{1:N} |\bm{x}_0)$, $\bm{x}_0=\bm{x}$, the diffusion process involves a Markov chain that incrementally introduces Gaussian noise to the data until its distribution approaches the latent distribution $\mathcal{N}(\bm{0}, \bm{I})$.
During training, instead of sequentially adding noise to the original data $\bm{x}_0$, we adopt the formulation proposed in \cite{ho2020denoising} to implement the diffusion process:
\begin{equation}
        \label{eq_add_noise_x0}
        q({x}_{t}^{1:N}\mid{x}_{0}^{1:N}) = \sqrt{\bar{\alpha_t} }{x}_{0}^{1:N} + \epsilon \sqrt{1-\bar{\alpha_t}}, \epsilon \sim \mathcal{N}(\bm{0},\bm{I}),
\end{equation}
where $\bar{\alpha_t} =  {\textstyle \prod_{i=0}^{t}\alpha_i}$. Thus, we can efficiently generate ${x}_t^{1:N}$ by sampling a noise value from a standard Gaussian distribution. From here we use $\bm{x}_t$ to denote the full sequence at noising step t. 
\paragraph{Reverse Conditional Gesture Generation.}
In our approach to conditioned gesture synthesis, we model the distribution $p_{\theta}(\bm{x}_{0}\mid \bm{c})$ using the reversed diffusion process of gradually removing noise from $\bm{x}_{T}$, given condition $\bm{c}$. Instead of directly predicting $\bm{x}_0$, we find that predicting the noise term $\epsilon$, i.e., $\epsilon_t = G(\bm{x}_{t}, t, \bm{c})$ leads to better performance.
For the task of co-audio gesture synthesis, we require additional inputs such as audio and speaker identity, which we treat as context information $\bm{c}$ and include as conditions in the generation process. To account for these conditions, we modify the formulation of the reverse process of each timestep as follows:
\begin{equation}
        p_{\theta }(\bm{x}_{t-1}\mid \bm{x}_{t},\bm{c})=\mathcal{N}(\bm{x}_{t-1}; \mu_{\theta }(\bm{x}_{t}, t, \bm{c}), {\textstyle \sum_{\theta }}(\bm{x}_{t}, t, \bm{c}) ). 
\end{equation}

\subsection{Integrating Emotion into Gesture}\label{sec:ieig}
\label{e2g}
There is currently limited research on generating emotive human gestures based on audio input. Previous approaches have involved using emotion labels alongside other input modalities (e.g. text) to generate emotive human gestures. In contrast, our method extracts emotional features directly from speech. 
Specifically, we first pool the audio features (introduced at Section \ref{sec:trans}) of length $N$ to $1$. This transforms a sequence audio features $\bm{A} \in \mathbb{R}^{N \times D_{a}}$  to $ \bm{\hat{A}}\in \mathbb{R}^{D_{a}}$. Suppose we have $C$ emotion categories. We apply a fully-connected layer $\phi$ that maps $\bm{\hat{A}}$ to the logits $c \in \mathbb{R}^{C}$, then we have the emotion category corresponding to the speech. 
We consider this emotional category to be a controlling factor in the generation of emotive gestures.

How to integrate emotion into gestures is crucial for achieving optimal performance. 
Several methods have been developed that enable the diffusion process to conditionally embed the additional input data, including adaptive normalization methods (e.g., AdaLN \cite{perez2018film} and AdaIN \cite{huang2017arbitrary}), in-context initialization \cite{peebles2022scalable,tevet2022human,zhu2023taming}, and attention-based conditioning \cite{rombach2022high,zhang2022motiondiffuse}. Here, we elaborate on the relationship between emotion and gesture sequences, providing a detailed exploration of methods for integrating emotion.

\textbf{Adaptive Normalization} leverage adaptive layer normalization or instance normalization to condition the diffusion process on a style vector, which can be used to control aspects of the output such as texture, color, or lighting. In this case, we envision a global relationship between emotion modality and gesture sequences, and we employ the Adaptive Layer Normalization (AdaLN) \cite{perez2018film} which modulates the channel-map distribution to apply the emotion modality to the entire gesture sequence. 
Formally, AdaLN learns functions $f$ and $h$ which output $\gamma$ and $\beta$ from the emotion embedding $\bm{e}$, and modulate into the gesture feature $\bm{x}_t$:
\begin{equation}
        \gamma = f(\bm{e}), \hspace{3mm}\beta = h(\bm{e}),
\end{equation}
\begin{equation}
        {\bm{x}}'_{t} = \gamma \cdot \bm{x}_{t} + \beta.
\end{equation}
The functions $f$ and $h$ are chosen without any specific criteria and, for this particular instance, we assign them to two distinct linear projections. Therefore, AdaLN modulates the channel-map distribution based on $\bm{e}$, agnostic to temporal location, achieving the global association between gesture and emotion.

\textbf{In-context Initialization} is another method establishing global connections and involves augmenting the noisy gesture sequence with an emotional modality, which is concatenated with the noisy gesture sequence and then fed into the model, enabling the model to condition its predictions on the emotional modality. The two primary ways of incorporating emotional input involve treating it either as a token or as content. With the former sending the emotion and noise gesture sequence as tokens to the transformer in the time dimension \cite{peebles2022scalable}, and the latter carrying out the integration in the channel dimension, both methods aim to integrate global emotional characteristics into the gesture sequence.

\textbf{Attention-based Conditioning} uses the emotional vector $\bm{e}$ to compute an attention score that modulates the vivid gesture generation at each frame, due to the envision that emotions are subject to temporal change, with their intensity varying from moment to moment.
In this setting, a commonly used cross-attention approach \cite{rombach2022high} is adopted. Suppose the gesture noise-level counterpart is $\mathbf{\bm{x}} \in \mathbb{R}^{N\times D_{a}}$ where $D_{a}$ is the feature dimension of gesture. In cross-modal cross-attention, the query ($\bm{Q}$) values are derived from gesture features, while the key ($\bm{K}$) and value ($\bm{V}$) matrices are obtained from the emotional vector. Formally,
\begin{equation}
\bm{Q} = W_{Q}\cdot \bm{x}_{t}, \hspace{2mm}
\bm{K} = W_{K}\cdot \bm{e}, \hspace{2mm} \bm{V} = W_{V}\cdot \bm{e}.
\end{equation}
Here, $t$ denotes the $t$-th time step in the diffusion process, and $W_{Q}, W_{K}, W_{V}$ are learnable projection matrices \cite{jahn2021high,yan2021videogpt}. Subsequently, the emotional middle representation $\bm{x}_{t}$ undergoes an update process:
\begin{equation}
\label{eq_cross_attention}
{\bm{x}}'_{t} = softmax(\frac{QK^{T}}{\sqrt{D_{a}} }) \cdot V.
\end{equation}

In addition, our approach enables spatial-level body part manipulation by adapting diffusion inpainting to gesture data. The editing process is performed exclusively during sampling and does not require any training. Specifically, we select a gesture sequence input, and during model sampling, we replace a portion of the joints in the gesture sequence with random noise $\sim \mathcal{N}(\bm{0}, \bm{I})$ and apply Equation \ref{eq_add_noise_x0} to add noise to the remaining joints, the experimental results are shown in Supplementary Material. 
This technique promotes coherence with the original input and facilitates the completion of missing parts in the generated output.

\subsection{Joint Correlation-aware Transformer (JCFormer)}\label{sec:trans}
Our methodology primarily emphasizes the generation of gestures through the relative skeletal space representation. This representation's inherent complexity makes learning gesture patterns using non-autoregressive (due to error accumulation) architectures challenging. Furthermore, the incorporation of audio as a driving factor for generating gestures, with the generated gestures being temporally reliant on the audio, further exacerbates the learning difficulties of the model. Therefore, we disentangle the gestures to joints correlation and inter-frame joints dynamics, and utilize the Transformer's robust ability in the modeling sequential data, unlike the majority of prior research which relies on recurrent networks \cite{liu2022learning,yoon2020speech}.
\paragraph{Audio-Gesture Alignment.} 
Our audio encoder is built on the state-of-the-art self-supervised pre-trained speech model, wav2vec 2.0 \cite{baevski2020wav2vec}. Following~\cite{fan2022faceformer}, we add a linear projection layer to the audio encoder, allowing for alignment with the gesture features that are sampled at a different frequency (e.g. $50$Hz for audio while $15$fps for gestures). Therefore, the resulting audio and gesture modalities are well aligned.

\paragraph{Joint Correlation-aware Transformer Module.}
Most recent methods in this field concentrate on extracting temporal features from input sequences. Nevertheless, we contend that spatial information, which encodes kinematic relationships between joints and correlations between different body parts, is as vital as temporal information. Therefore, a joint correlation-aware Transformer module is designed to fuse the spatial-joint and temporal information. As illustrated in Figure \ref{fig:network}, our JCFormer consists of two core components: the joint-aware transformer and the temporal-aware transformer. 

The aim of the joint-aware transformer is to capture both global interdependencies between various body parts as well as local interdependencies between different joints. To preserve both global spatial-joint correlations (e.g., moving the head is often related to moving the shoulders and torso) and local spatial-joint correlations (e.g., the movements of the thumb and index finger in a grasping gesture) across different frames, we introduce an additional learnable correlation token to the sequence. The state of this token at the output of the joint-aware transformer encoder serves as the spatial representation, similar to the classification token proposed in ViT \cite{dosovitskiy2020image}. Given an input noise-level gesture sequence $\bm{x}_{t} \in \mathbb{R}^{N \times (J\cdot 3)}$, we reshape it to $\bm{x}_{t}^{s} \in \mathbb{R}^{N \times J\times 3}$, where $N$ is the number of frames, $J$ is the number of joints, and $3$ represents the 3D coordinates of each joint. Then, the temporal dimension is eliminated by applying a linear transformation $\in \mathbb{R}^{N\times 1}$.
We employ the temporal-aware transformer to obtain the $\hat{\bm{x}}_{t}$, which have temporal dynamics of gesture, from an input noise-level gesture sequence $\bm{x}_{t}$.
Finally, the correlation token is broadcast to the $\hat{\bm{x}}_{t}$ for fusion processing.

\paragraph{Audio to Gesture Cross-Attention.} 
The choice of which method to utilize for the integration of additional modalities in diffusion models is contingent upon the interrelationships among different modalities. Therefore, careful consideration of the nature and characteristics of the input data is crucial to determine the appropriate conditioning method. In our task, the temporal alignment between audio and gesture is highly pronounced, rendering cross-attention an ideal choice for their fusion. Suppose the audio feature extracted by the audio encoder is $\bm{A} \in \mathbb{R}^{N\times D_{a}}$, where $D_{a}$ represents the feature dimension of audio. In this cross-attention, the query values are derived from gesture features, while the key  and value matrices are obtained from audio features. The calculation of cross-modal cross-attention (the same as Equation \ref{eq_cross_attention}) involves the conversion of audio features to gesture features, facilitating the temporal alignment and fusion of the two modalities. 

\subsection{Training Objective}
During the gesture generation training process, at each iteration timestep of the reverse diffusion process, $\hat{\bm{x}}_{0}$ is calculated from the predicted noise $\hat{\epsilon}_{t}$ and intermediate noise variable $\bm{x}_{t}$:
\begin{equation}
        \hat{\bm{x}}_{0} = \frac{\bm{x}_{t}-\sqrt{1-\bar{\alpha}_{t}} \hat{\epsilon}_{t} }{\sqrt{\bar{\alpha}_{t} } }.
\end{equation}

Then, we introduce a geometric reconstruction loss, which encourages the generated gesture $\boldsymbol{\hat{x}}_0=\{\hat{\bm{x}}^{i}_{0}\}_{i=1}^N$ to match the ground-truth gesture as closely as possible:
\begin{equation}
        L_{rec} = {E}_{i \sim [1, N] } \left \| \bm{x}_{0}^{i} - \hat{\bm{x}}^{i}_{0}  \right \| _{2}.
\end{equation}

Additionally, to optimize the diffusion model, we use the denoising diffusion objective \cite{nichol2021glide}:
\begin{equation}
        L_{mse} = E_{\epsilon \sim \mathcal{N}(\bm{0}, \bm{I}), t\sim [1, T]}\left [ \left \| \epsilon -G(\bm{x}_{t}, t, \bm{c}) \right \| _{2} \right ].
\end{equation}

We utilize a cross-entropy loss to optimize the alignment between the predicted probability and the true emotion category $l$:
\begin{equation}
        L_{ce} = CrossEntropy(c, l).
\end{equation}

Overall, the optimization objective in our approach is defined as follows:
\begin{equation}
        L = L_{mse} + \lambda_{rec} \cdot L_{rec} + L_{ce}. 
\end{equation}

\section{Experiments}
\subsection{Experimental Setup}
\paragraph{Co-speech Emotional Dataset.}
As a large-scale semantic and emotional multi-modal dataset for conversational gestures synthesis, BEAT dataset \cite{liu2022beat} contains 76 hours of multi-modal data captured from 30 speakers talking with eight different emotions and four different languages. Following the benchmark \cite{liu2022beat}, we only utilize the speech data of English speakers, which amounts to approximately 16 hours in total. To enhance the model's timing robustness, we adopt a variable-length training strategy that involves randomly applying a proportional mask on the input sequence. To ensure continuity of the generated gestures during synthesis, we incorporated a 4-frame overlap between each clip for consecutive syntheses.
\paragraph{Comparison Methods.}
We compare our method with four state-of-the-art approaches: \textit{Speech2Gesture} \cite{ginosar2019learning} generates speech gestures using spectrograms as input, employing an encoder-decoder neural architecture with adversarial training. \textit{Trimodal} \cite{yoon2020speech} incorporates a multimodal context and employs adversarial training for automatic gesture generation. \textit{CaMN} \cite{liu2022beat} utilizes a cascaded architecture with multiple inputs, including audio, transcript, emotion labels, speaker ID, and facial blendshape weights, to generate gestures. \textit{DiffGesture} \cite{zhu2023taming} utilizes a diffusion-based co-speech gesture framework but did not converge in our experiments with the same settings. This may be attributed to the challenges posed by the relative-skeletal space representation used in the BEAT dataset. Consequently, we exclude the results of DiffGesture\cite{zhu2023taming} from Table \ref{tab-quantitative}.

\paragraph{Implementation Details.}
For a fair comparison, we set $N=34$ frame clips with stride 10, $J=47$ represent upper body joints for training for all the methods, and the variable-length training strategy $S_{vl}$ aforementioned where we applied a sliding window of 150 frames with a 50-frame step size, is treated as an additional variant for comparison. For JCFormer, we set $L$ = 8, $L_s$ = 4. The latent dimension of the joint-aware transformer, temporal-aware transformer, and fusion transformer is 64, 512, and 512 respectively. The latent dimension of the audio feature is 128. As for the diffusion process, the diffusion step is 1000. A batch size of 128 was used for training on a single NVIDIA A100 GPU.

\subsection{Evaluation Metrics}
For evaluation, we employ three metrics commonly used in co-speech gesture generation and related fields \cite{yoon2020speech,liu2022beat,zhang2022motiondiffuse}.
\paragraph{Fr$\acute{\boldsymbol{e}}$chet Gesture Distance (FGD).} FGD \cite{yoon2020speech} is a metric used to measure the distance between the distribution of synthesized gestures and real data, similar to FID in image generation studies, and is considered to be the most effective metric used in co-speech gesture generation evaluation. FGD is defined as the Fr$\acute{\boldsymbol{e}}$chet distance between the Gaussian mean and covariance of the latent features of human gestures X and those of the generated gestures $\hat{X}$:
\begin{equation}
        FGD(X, \hat{X}) = \left \| \mu _{r} -\mu_{g} \right \| ^{2} + Tr(\Sigma_r + \Sigma_{g} - 2(\Sigma_r \Sigma_g)^{1/2}),
\end{equation}
where $\mu_r$ and $\Sigma_r$ are the first and second moments of the latent feature distribution $Z_r$ of real human gestures $X$, while $\mu_r$ and $\Sigma_r$ come from the generated gestures.

\paragraph{Semantic-Relevant Gesture Recall (SRGR)}  \cite{liu2022beat} utilizes the semantic scores as weights to calculate the Probability of Correct Keypoint (PCK), which is the number of joints successfully recalled against a specified threshold $\delta$.

\paragraph{Beat Alignment Score (BeatAlign).} BeatAlign~\cite{li2021ai} evaluates the gesture-audio correlation by measuring the similarity between the kinematic beats and audio beats. BeatAlign is then defined as the average Chamfer distance between each kinematic beat and its nearest audio beat: 
\begin{equation}
        BeatAlign = \frac{1}{n}\sum_{i=1}^{n} exp(-\frac{min_{\forall b_{j}^{a}\in B_{a}}\left \| b_{i}^{m} - b_{j}^{a} \right \|^{2} }{2\sigma ^{2}} ),
\end{equation}
where $B^{m}=$\{$b_{i}^{m}$\} is the kinematic beats, $B^{a}=$\{$b_{j}^{a}$\} is the audio beat, and $\sigma$ is a parameter to normalize sequences, $\sigma = 0.3$ empirically.

\begin{table}[t]
\caption{\textbf{Quantitative Results}. To ensure a fair comparison, we employ the EMoG without variable-length training strategy ($S_{vl}$) as a variant to compare with other methods and present the average value of each metric obtained from running 10 times on the test data. }
\centering
\label{tab-quantitative}
\fontsize{9}{12}\selectfont
\setlength{\tabcolsep}{3mm}
\begin{tabular}{lccc}
    \toprule
    $Method$ & $FGD$ $\downarrow$     & $SRGR$ $\uparrow$  & $BeatAlign$ $\uparrow$ \\
    \midrule
    Speech2Gesture\cite{ginosar2019learning}                 & 256.7           & 0.092       & 0.751  \\
    MultiContext\cite{yoon2020speech}        & 177.2           & 0.227       & 0.751  \\
    CaMN\cite{liu2022beat}          & 122.8           & 0.240       & 0.783  \\ \hline
    \textbf{EMoG w/o $S_{vl}$ (ours)}           & \textbf{55.79}  & \textbf{0.195}  & \textbf{0.915} \\
    \textbf{EMoG (ours)}         & \textbf{48.26}      & \textbf{0.215}  & \textbf{0.922} \\
    \bottomrule
\end{tabular}
\end{table}

\subsection{Comparison Results}
\textbf{Quantitative Comparison.} The quantitative comparison results are reported in Table \ref{tab-quantitative}. We can find that the proposed EMoG demonstrates superior evaluation results across most metrics, especially outperforming existing methods by a large margin on FGD and BeatAlign scores, since our method utilizes emotion attributes to enhance joint distribution modeling of audio and gesture, enabling high-fidelity gesture reconstruction through the decoupling of spatial-joint relationships and temporal dynamics. The diversity introduced by the one-to-many mapping of audio to gestures and the sampling process of the diffusion model can adversely affect the SRGR metric, since the generated gestures may not always be close to the ground truth.

\begin{figure*}[t]
	\centering
	\includegraphics[width=1.0\textwidth]{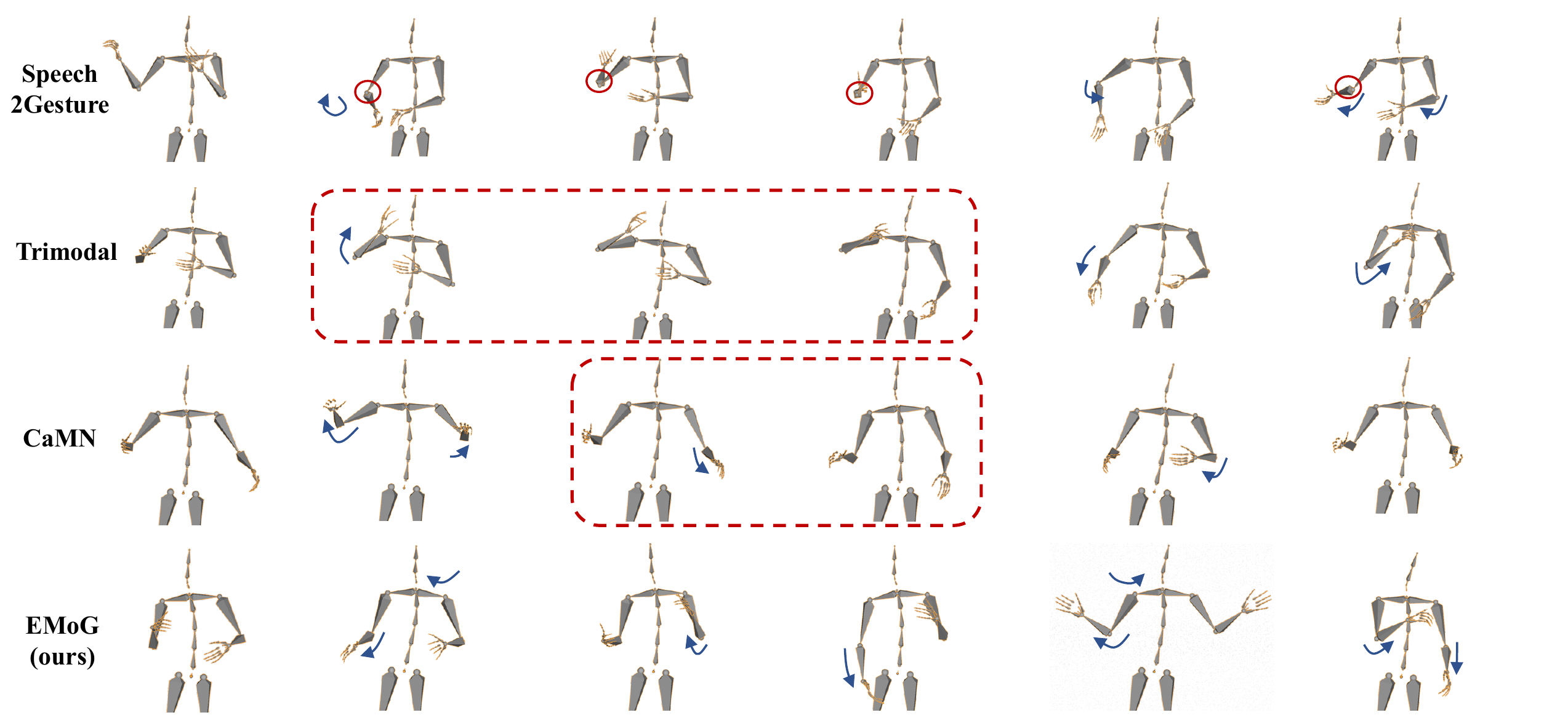}
	\caption{Qualtative Results on the BEAT dataset. We utilize circles to highlight incorrect joint orientations, while dull gestures are emphasized using rectangles. The direction of movement is indicated by the blue arrow.}
	\label{fig:compare_with_SOTAs}

\end{figure*}

\textbf{Qualitative Comparison.} We present the visualization results of various methods in Figure \ref{fig:compare_with_SOTAs}, showcasing selected keyframes with a time interval of 1.3 seconds (20 frames with 15fps). Since the relative skeletal space representation is used in BEAT dataset, there are certain requirements for the generation ability of the model. Comparative methods frequently generate slow and rigid poses, and Speech2Gesture \cite{ginosar2019learning}, which has a relatively simple encoder-decoder architecture, exhibits a tendency to predict incorrect joint orientations, such as the right hand being folded backward. In contrast, Emo can generate diverse and realistic gestures that are aligned with the corresponding speech audio.

\subsection{Emotion Integration}

In our experimental investigation of integrating emotion into gesture, we explore various approaches (described in Section \ref{e2g}). As shown in Table \ref{tab:tab-e2g}, our findings reveal that employing adaLN to modulate the channel-map distribution and using the emotion modality to the entire gesture sequence achieves the most optimal modeling of the joint distribution, outperforming other methods. The attention-based conditioning method also yields favorable results when incorporating emotions. However, the in-context initialization method is not well-suited for integrating emotions into gesture generation, as the model struggles to extract meaningful emotional distinctions from the context. With the emotion attribute extracted from the audio, EMoG can generate vivid gestures with enrichment $Fear$ emotion, compared to the EMoG w/o emotion attribute variants as shown in Figure \ref{fig:emotion_transfer}.

\subsection{Emotion Transfer}
\begin{figure*}[t]
	\centering
	\includegraphics[width=1.1\textwidth]{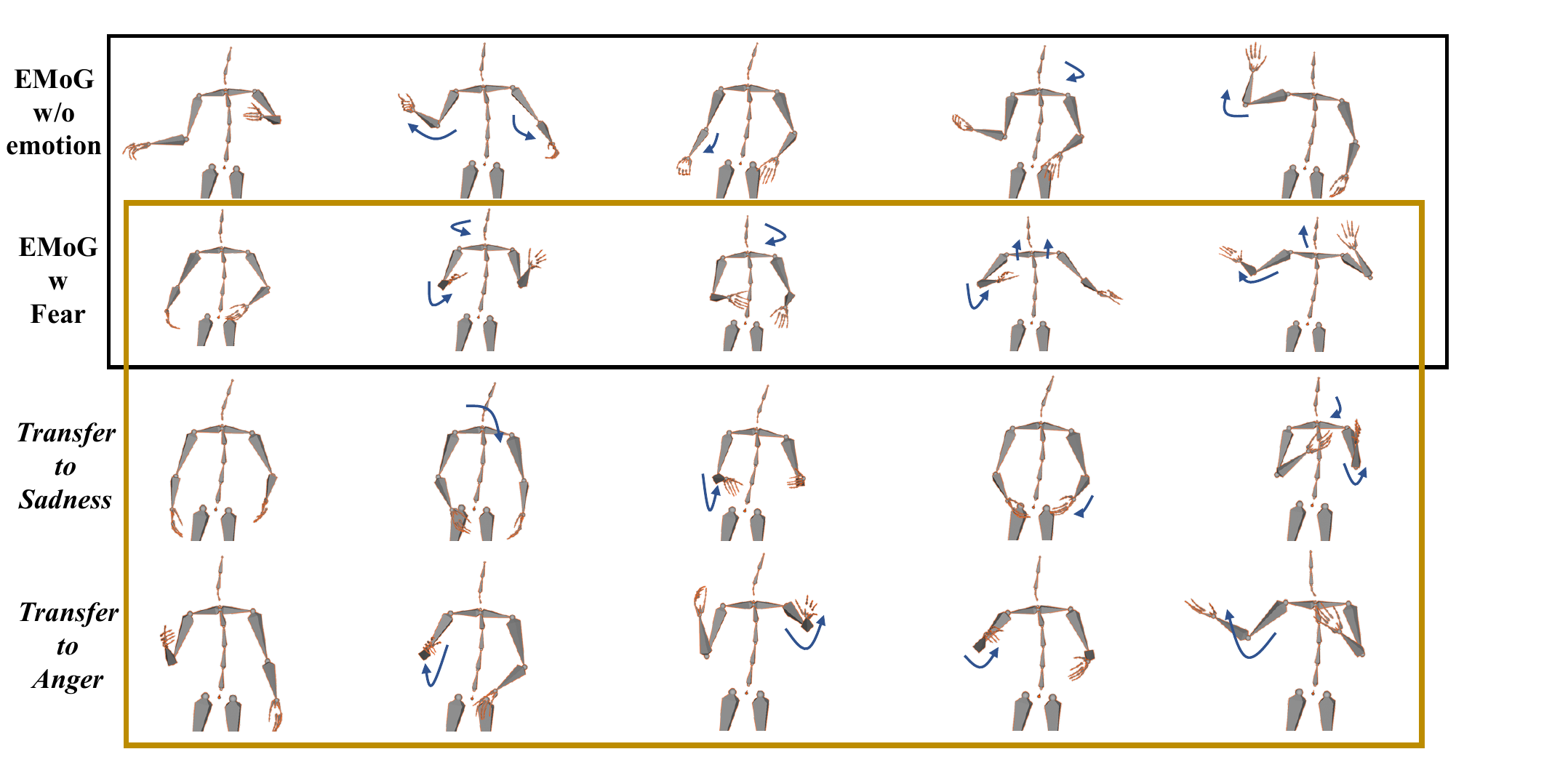}
	\caption{{Emotive Gesture Results. We compare the visual results between EMoG and the EMoG w/o emotion attribute variants in $black$ rectangle. The emotive gesture results from the $Fear$ emotion attribute transfer to $Sadness$ and $Anger$, are shown in $yelllow$ rectangle. The direction of gesture movement is indicated by the blue arrow.}}
	\label{fig:emotion_transfer}
\end{figure*}

Emotion transfer to the same audio-generated gesture is accomplished by modifying the embedded emotion attribute. Since the emotion attribute extracted from audio is supervised by the ground truth emotion label, we can achieve emotion transfer of gestures by providing the corresponding emotion label. To illustrate this, we present the results in Figure \ref{fig:emotion_transfer}. Given audio, we transfer the emotion attribute from $Fear$ to $Sadness$ and $Anger$. We found that $Fear$ gestures are characterized by iconic shrugging and dodging, $Sadness$ gestures tend to change gradually without exaggerated gestures, and head posture is always downcast, whereas $Anger$ gestures exhibit rapid and violent changes.

\subsection{Ablation Studies}
In this part, we will demonstrate the effectiveness of our proposed EMoG framework by conducting ablation studies on key modules. Specifically, we will analyze the impact of removing the reconstruction loss $L_{rec}$, the JCFormer, and the emotion attribute $Emotion$ on performance, as reflected in Table \ref{tab-ablation}. We discovered a significant decrease in the FGD metric when the emotion attribute was removed, providing evidence that the extracted emotion attribute from audio effectively models the joint distribution of audio and gesture. After the removal of the joint correlation-aware Transformer, there is a substantial decrease in the SRGR metric, suggesting that joint-spatial expression plays a role in generating high-quality gestures. Furthermore, it is evident that the additional reconstruction loss also has an impact on the quality of gesture generation. 

\begin{table*}
\begin{floatrow}
\capbtabbox{
\setlength{\tabcolsep}{1mm}
\begin{tabular}{lcc}
    \toprule
    Method                          & $FGD$ $\downarrow$  \\
    \midrule
    In-context initialization       & 61.25               \\
    Attention-based conditioning    & 53.61               \\
    AdaLN                           & \textbf{48.26}       \\
    \bottomrule
\end{tabular}
}{
 \caption{Results using different methods to integrate emotion into gesture. }
 \label{tab:tab-e2g}
}
\capbtabbox{
\setlength{\tabcolsep}{1mm}
\begin{tabular}{ccccc}
\toprule
 $L_{rec}$ & JCFormer   & $Emotion$  & $FGD$ $\downarrow$   & $SRGR$ $\uparrow$    \\ \midrule
           &\checkmark          &            &  66.18  &0.196    \\
 \checkmark & \checkmark         &            & 61.44 &0.204      \\
 \checkmark&                    & \checkmark &  54.51  &0.201  \\
 \checkmark& \checkmark          &\checkmark  & \textbf{48.26}  &\textbf{0.215}     \\ \bottomrule
\end{tabular}
}{
 \caption{Quantitative results on ablation study.}
 \label{tab-ablation}
}
\end{floatrow}
\end{table*}

\vspace{-10pt}
\section{Conclusion}
This study presents a novel framework, EMoG, which leverages denoising diffusion models to synthesize emotive co-speech 3D gestures. To infuse vividness into the synthesized gesture, our proposed method seamlessly integrates emotional representations extracted from audio into the distribution modeling process of the diffusion model. Furthermore, to improve the realism and accuracy of the generated gesture, we introduce a joint correlation-aware transformer capable of simultaneously predicting the skeletal modeling and the temporal dynamics of the joints.
While EMoG demonstrates promising performance in the co-speech gesture generation task, it still encounters certain challenges. For instance, during the extraction of emotional attributes, supervision through emotion labels is still necessary for the training phase. Our future work aims to explore the spontaneous extraction of additional properties (e.g. emotion) from audio to enhance the generation of more vivid synchronized speaking gestures.

\medskip

\bibliographystyle{unsrtnat}
\bibliography{EMoG}

\section{Supplementary Material}

\subsection{Distinctive Emotion Gesture Generation}

\subsubsection{Emotional Gestures Guided by Different Audio }
The audio feature encompasses emotionally relevant information that is expected to affect the variations in gestures throughout a global timeframe. In our method, the emotion attribute is directly extracted from audio, which can be considered a controlling factor (this branch is more like Speech Emotion Recognition (SER) task). As illustrated in Figure \ref{fig:t-SNE} (a), Our method effectively discriminates between emotions associated with different audio inputs, allowing for the generation of distinct emotional gestures. Indeed, this implies that we can generate emotionally expressive gestures without requiring additional input of emotion labels. The emotional information embedded within the audio feature is sufficient for generating gestures that convey the desired emotional expression.
\subsubsection{Distinctive Emotion-specific Gestures Guided by the Same Audio}
\begin{figure*}[t]
	\centering
        \includegraphics[scale=0.54]{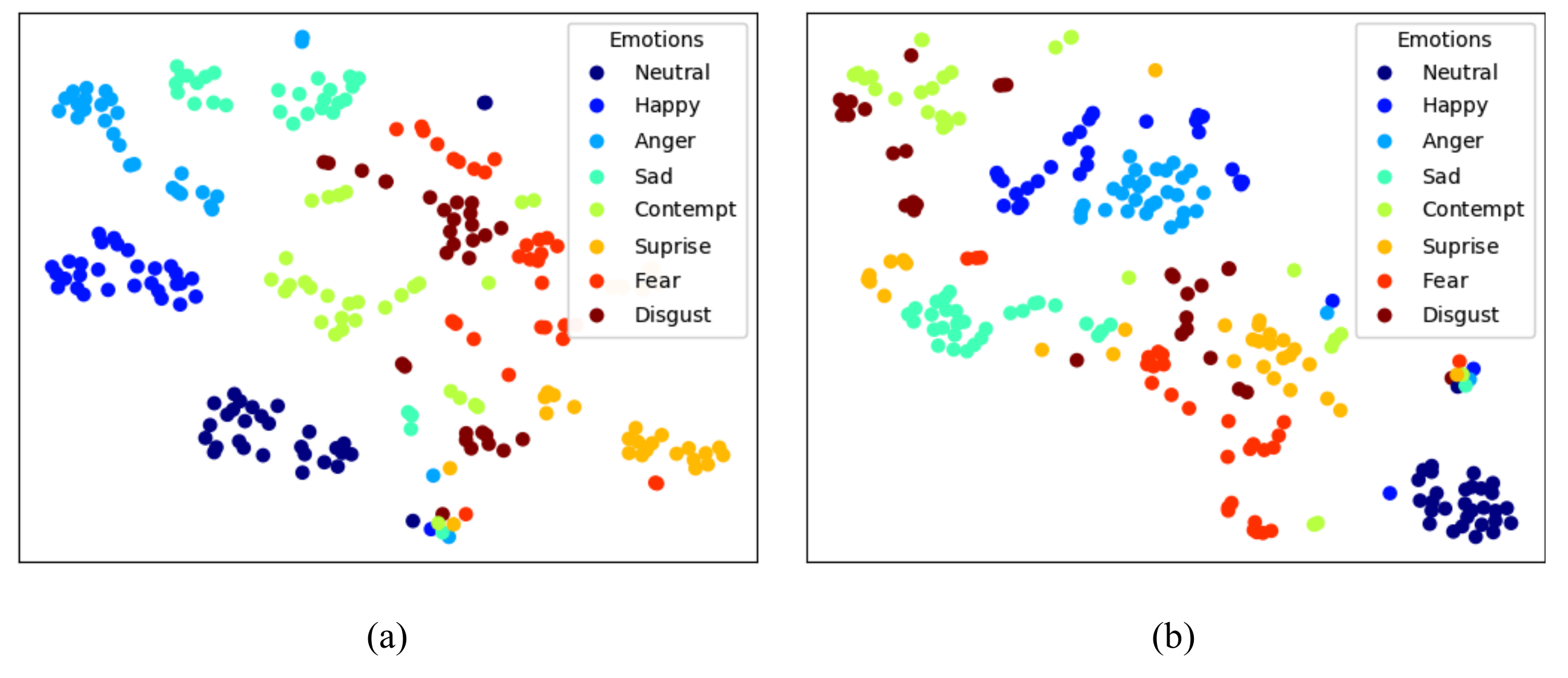}
	\caption{The t-SNE visualization of generated gestures. (a) We randomly choose 30 generated gesture clips per emotional audio (total eight emotions). Results show that our method effectively discriminates between emotions associated with different audio inputs, allowing for the generation of distinct emotional gestures. (b) Given the same audio, EMoG enables precise emotional control over the generated gestures.}
	\label{fig:t-SNE}
\end{figure*}

Additionally, we offer a preset emotion control interface that allows users to specify desired emotions. Through this interface, users can generate gestures with specific emotions according to their preferences. Given an audio input, our method has the capability to generate emotion-specific gestures by incorporating different emotions as input. This enables us to generate gestures that align with and convey the desired emotional states.
We choose 30 audio segments to generate eight emotive gestures respectively, the t-SNE visualization of generated emotive gestures is shown in Figure \ref{fig:t-SNE} (b). 
By providing emotional input, our method can generate distinctive emotion-specific gestures that accurately reflect the intended emotions.

\subsection{Fine-grained Body Part Controllable Gesture Generation}
\begin{figure*}[t]
	\centering
        \includegraphics[scale=0.43]{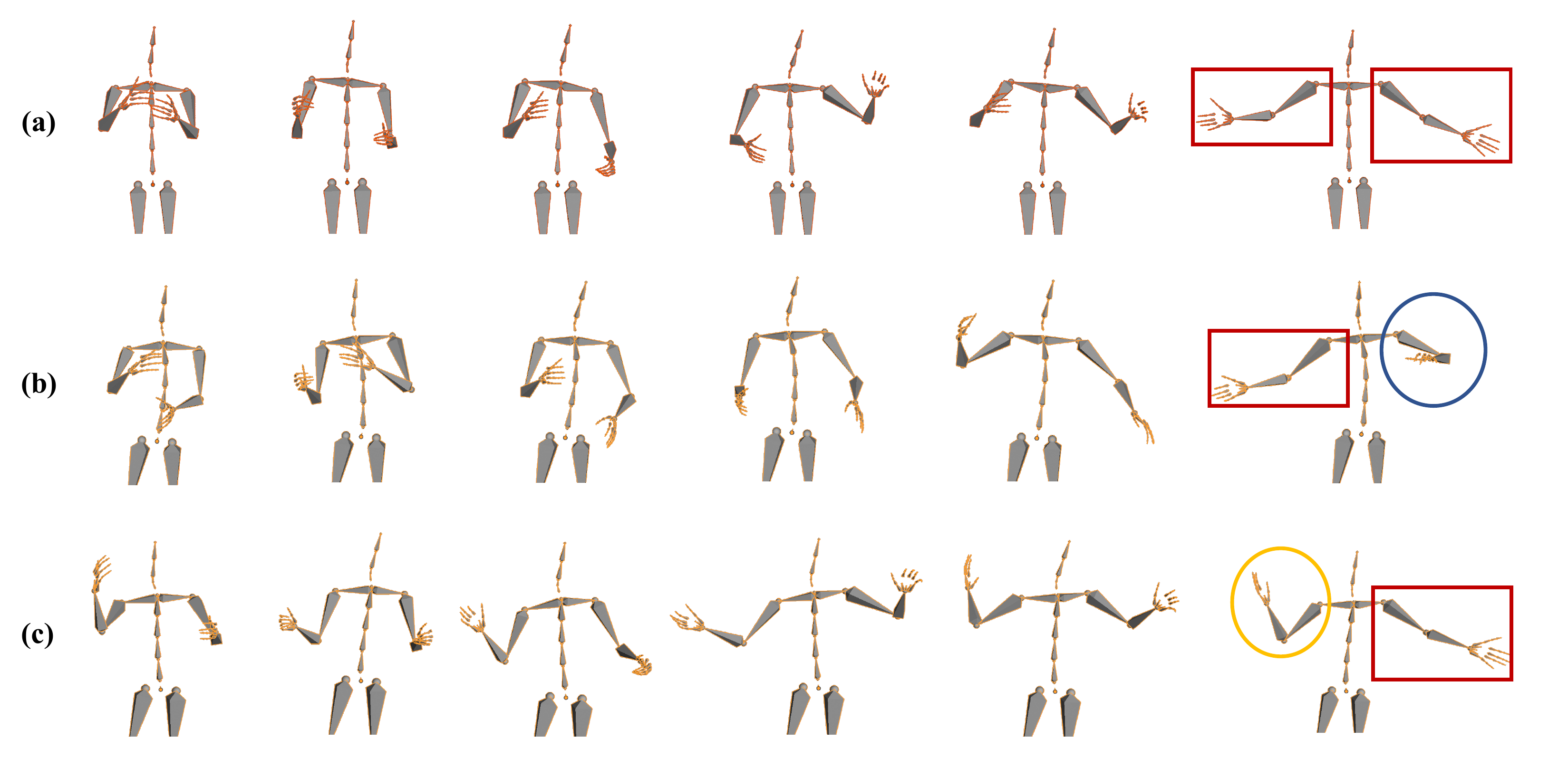}
	\caption{The visualization of controllable hand gestures. (a) Reference motion. (b) generated controllable left-hand gestures. (c) generated controllable right-hand gestures. Red rectangles indicate invariant hand movements, blue circles represent left-hand motion generation, and yellow circles represent right-hand motion generation.}
	\label{fig:control_hands}
\end{figure*}

Our method enables fine-grained control over specific body parts for emotion generation. This allows for precise manipulation and generation of emotions in relation to different body parts, resulting in detailed and controllable gestures. 
Specifically, we have the ability to selectively generate motions for specific body parts, such as the left and right hands. To achieve this, we replace the joints of the targeted body parts with standard Gaussian noise, while adding noise of T time steps to the remaining body parts (as described in Equation \ref{eq_add_noise_x0}). This method allows us to generate gestures that smoothly control the desired parts while maintaining the stability of other parts that do not require significant changes. The effectiveness of this approach is demonstrated in Figure \ref{fig:control_hands}.

\subsection{Network Architecture Details}
In this section, we present a more specific EMoG architecture. 
\subsubsection{Audio Encoder}
Our audio encoder is built on the state-of-the-art self-supervised pre-trained speech model, wav2vec 2.0 \cite{baevski2020wav2vec}, consisting of an audio feature extractor and a multi-layer transformer encoder \cite{vaswani2017attention}. The audio feature extractor, which is composed of temporal convolution layers (TCN), converts the raw waveform input into feature vectors with a frequency of $f_a$. The transformer encoder consists of multi-head self-attention and feed-forward layers, which contextualize the audio feature vectors and generate speech representations. The quantization module discretizes the temporal convolution outputs into a finite set of speech units. Similar to masked language modeling \cite{devlin2018bert}, wav2vec 2.0 solves a contrastive task to identify the true quantized speech unit by utilizing the context surrounding a masked time step. Following~\cite{fan2022faceformer}, we add a linear projection layer to the audio encoder, allowing for alignment with the gesture features that are sampled at a different frequency (e.g. $50$Hz for audio while $15$fps for gestures). Therefore, the resulting audio and gesture modalities are well aligned.

\subsubsection{Joint Correlation-aware Transformer Module}
\paragraph{Positional Encoding.} Since our model is based on the Transformer architecture, it is necessary to introduce information regarding the relative or absolute position of the markers in the sequence. This allows the model to leverage the order of the sequence effectively. Since the skeleton exhibits a relatively fixed structure, we employ the sinusoidal absolute positional encoding technique \cite{vaswani2017attention} in the joint-aware transformer. This encoding scheme enables the joint-aware transformer to capture and utilize the positional information of the joints effectively. Since the temporal relationship of gesture sequences is more complex, we utilize a learnable positional encoding method \cite{zhang2022motiondiffuse} to capture the temporal dependencies within the gestures. This learnable positional encoding method enhances the temporal-aware transformer's ability to effectively capture and model the sequential dynamics of the gestures over time.

\begin{figure*}[t]
	\centering
        \includegraphics[scale=0.7]{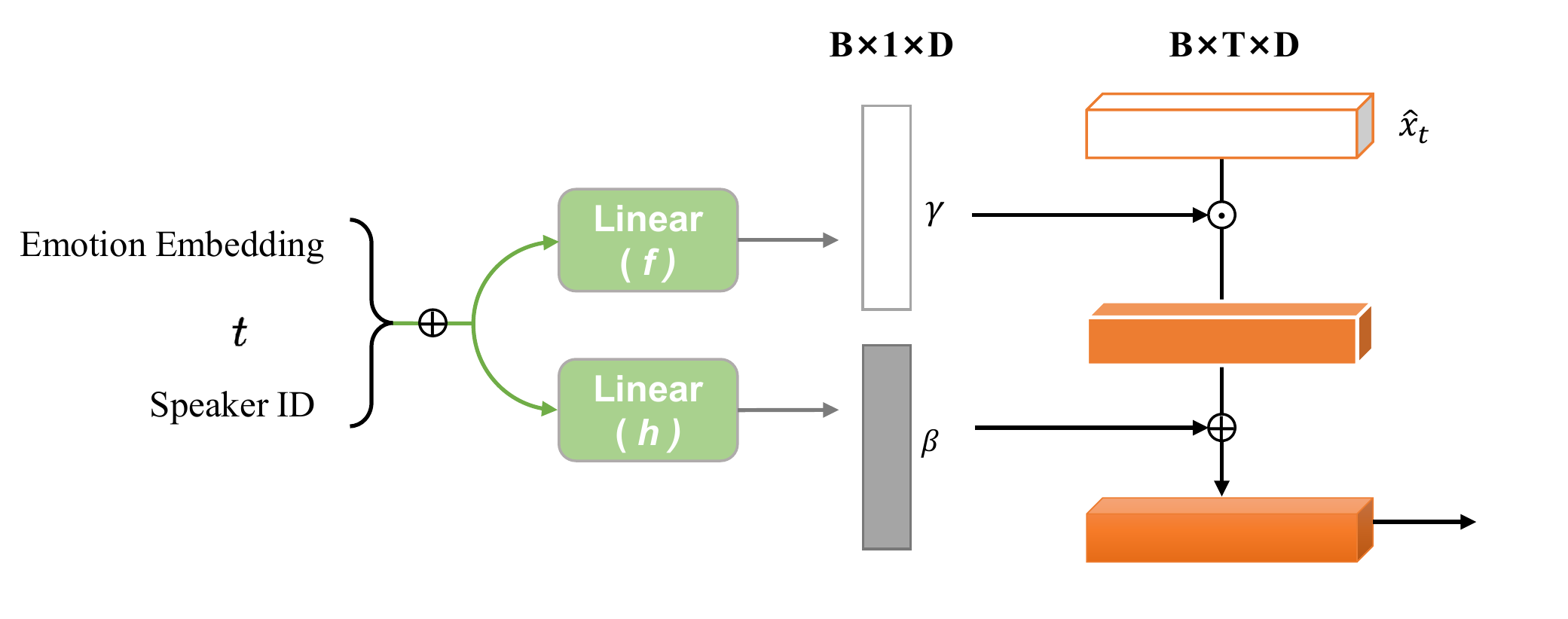}
	\caption{A single AdaLN layer. The dot signifies a Hadamaed product, which is an element-wise multiplication between two matrices.}
	\label{fig:adaln}
\end{figure*}

\paragraph{Feature-wise Linear Style Modulation} As discribaled in Section \ref{sec:ieig}, we leverage the AdaLN \cite{perez2018film} to modulates the channel-map distribution on emotion atrribute. We found this method to be particularly beneficial for emotional embedding. For the two additional inputs, the timestamp t and speaker ID, we employ the same embedding technique to incorporate them into the model, the detailed implementation is illustrated in Figure \ref{fig:adaln}.

\subsubsection{Integrating Seed Pose in Gesture Generation}
Given a seed pose consisting of several frames, generating gestures along the seed pose is a common approach in audio-driven gesture generation tasks. The purpose is to generate gestures in a manner that follows the temporal dynamics of the seed pose, resulting in smoother motion sequences in the time dimension \cite{yoon2020speech}. Previous methods often incorporate the seed pose as one of the inputs during the training process, which can restrict the practical application of the method in industrial settings. Hence, we propose EMoG, a model that does not require an additional seed pose input during the training phase. During the inference phase, the seed pose can be optionally employed as an additional conditional input. This design allows for more flexible application of the EMoG, accommodating both scenarios where the seed pose is available and where it is not necessary. 
In the specific case of incorporating the seed pose, we replace the initial 4 frames of random noise from a standard Gaussian distribution with the seed pose, augmented by T time steps noise counterparts (as described in Equation \ref{eq_add_noise_x0}). 
By adopting this approach, the distribution of generated gestures can be further adjusted based on the provided seed pose.

\medskip

\subsection{Details of BEAT Dataset}
\begin{figure*}[t]
	\centering
        \includegraphics[scale=0.45]{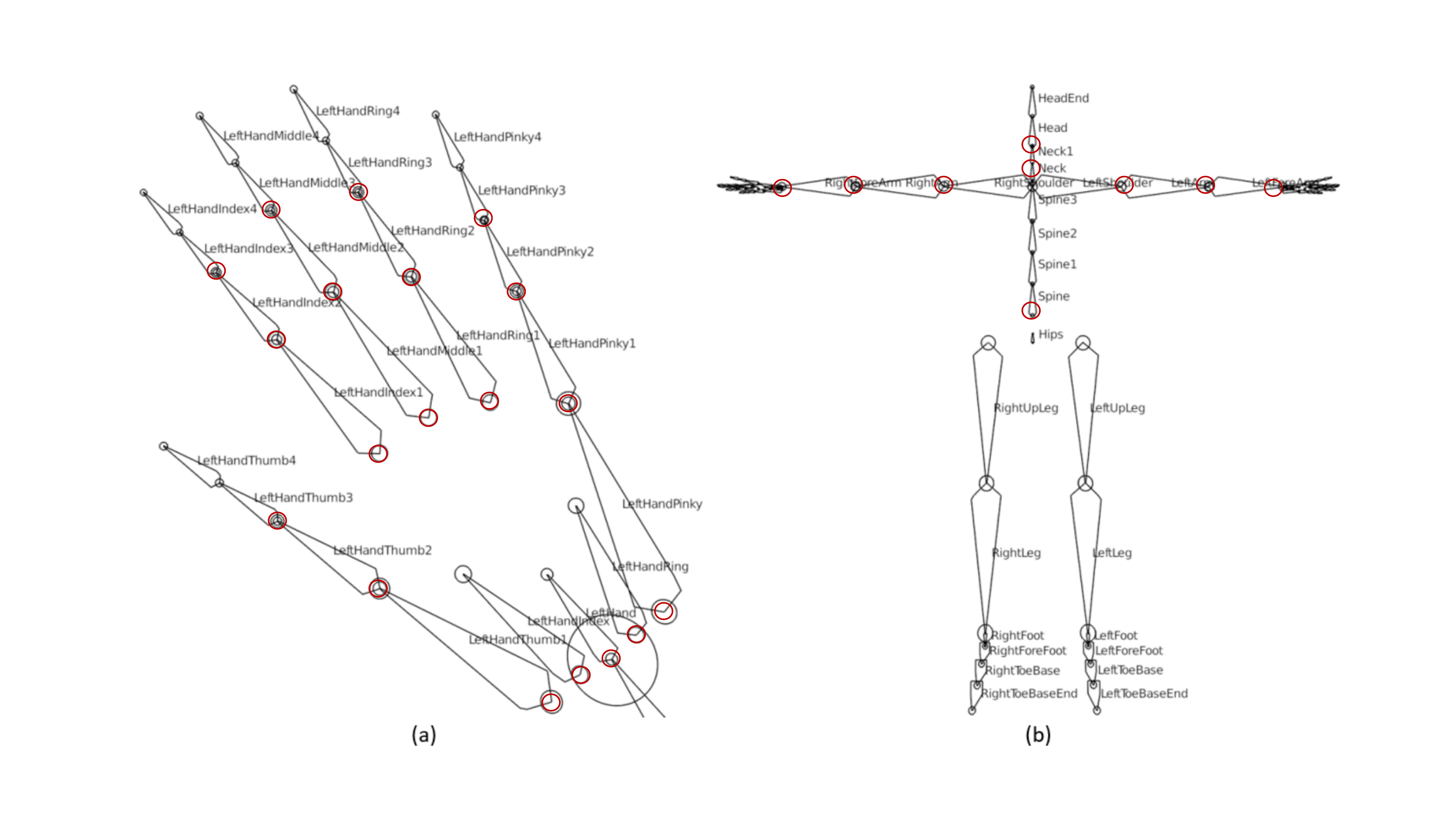}
	\caption{The skeleton of body and hands. The highlighted part represents the joint that is utilized by the EMoG model, which contain 38 hand joints (a) and 9 body joints (b).}
	\label{fig:skeleton}
\end{figure*}

BEAT \cite{liu2022beat} is a large-scale, high-quality, multi-modal dataset that incorporates semantic and emotional annotations. The dataset consists of eight distinct emotional annotations, including neutral, happiness, anger, sadness, contempt, surprise, fear, and disgust. The human skeleton on the BEAT dataset has 48 hand joints and 27 body joints. Follow the offical set \cite{liu2022beat}, we only utilize specific joints located in the upper body for our synthesis, as shown in Figure \ref{fig:skeleton}. 

\end{document}